\documentclass[10pt,twocolumn,letterpaper]{article}
\PassOptionsToPackage{table,xcdraw}{xcolor}
\usepackage[pagenumbers]{cvpr} 

\definecolor{cvprblue}{rgb}{0.21,0.49,0.74}
\usepackage[pagebackref,breaklinks,colorlinks,allcolors=cvprblue]{hyperref}
\usepackage{multirow}
\usepackage[dvipsnames]{xcolor}
\usepackage{color}
\usepackage{CJKutf8}

\def\paperID{1881} 
\def\confName{CVPR}
\def\confYear{2025}

\title{DSPNet: Dual-vision Scene Perception for Robust 3D Question Answering}

\author{Jingzhou Luo$^1$ \hspace{0.1em} Yang Liu$^1$\thanks{Corresponding Author} \hspace{0.1em}  Weixing Chen$^1$ \hspace{0.1em}  Zhen Li$^2$ \hspace{0.1em} Yaowei Wang$^3$ \hspace{0.1em} Guanbin Li \hspace{0.1em} Liang Lin$^1$\\
$^1$Sun Yat-sen University $^2$The Chinese University of Hong Kong $^3$Peng Cheng Laboratory\\
{\tt\small \{luojzh5,chenwx228\}@mail2.sysu.edu.cn, \{liuy856,liguanbin\}@mail.sysu.edu.cn,}\\
{\tt\small lizhen@cuhk.edu.cn,  wangyw@pcl.ac.cn, linliang@ieee.org}
}

\begin{document}
\maketitle
\begin{abstract}
 3D Question Answering (3D QA) requires the model to comprehensively understand its situated 3D scene described by the text, then reason about its surrounding environment and answer a question under that situation. However, existing methods usually rely on global scene perception from pure 3D point clouds and overlook the importance of rich local texture details from multi-view images. Moreover, due to the inherent noise in camera poses and complex occlusions, there exists significant feature degradation and reduced feature robustness problems when aligning 3D point cloud with multi-view images. In this paper, we propose a  \textbf{Dual-vision Scene Perception Network (DSPNet)}, to comprehensively integrate multi-view and point cloud features to improve robustness in 3D QA. Our Text-guided Multi-view Fusion (TGMF) module prioritizes image views that closely match the semantic content of the text. 
 To adaptively fuse back-projected multi-view images with point cloud features, we design the Adaptive Dual-vision Perception (ADVP) module, enhancing 3D scene comprehension. Additionally, our Multimodal Context-guided Reasoning (MCGR) module facilitates robust reasoning by integrating contextual information across visual and linguistic modalities. Experimental results on SQA3D and ScanQA datasets demonstrate the superiority of our DSPNet. Codes will be available at \href{https://github.com/LZ-CH/DSPNet}{https://github.com/LZ-CH/DSPNet}. 
 

\end{abstract}    
\section{Introduction}
\label{sec:intro}

Recently, 3D Question Answering (3D QA), the task of answering questions about 3D scenes, has emerged as a significant research area in artificial intelligence~\cite{liu2024aligning}. Unlike traditional 2D Question Answering (2D QA), which relies on flat images, 3D QA offers the potential for richer spatial comprehension and immersive interaction. The expansion of QA tasks from 2D to 3D also broadens the scope of cross-domain applications, such as visual language navigation ~\cite{anderson2018vision, wang2019reinforced}, embodied agents ~\cite{savva2019habitat, huang2023leo} , and autonomous driving ~\cite{qian2024nuscenes}. However, compared with the 2D VQA task, 3D QA introduces unique challenges that extend beyond planar visual understanding. These challenges primarily manifest in the necessity to accurately perceive complex geometric relations among multiple scene entities and effectively reason about spatially semantic dependencies through natural language interactions in 3D environments. 
\begin{figure}
    \centering
    \includegraphics[width=1.0\linewidth]{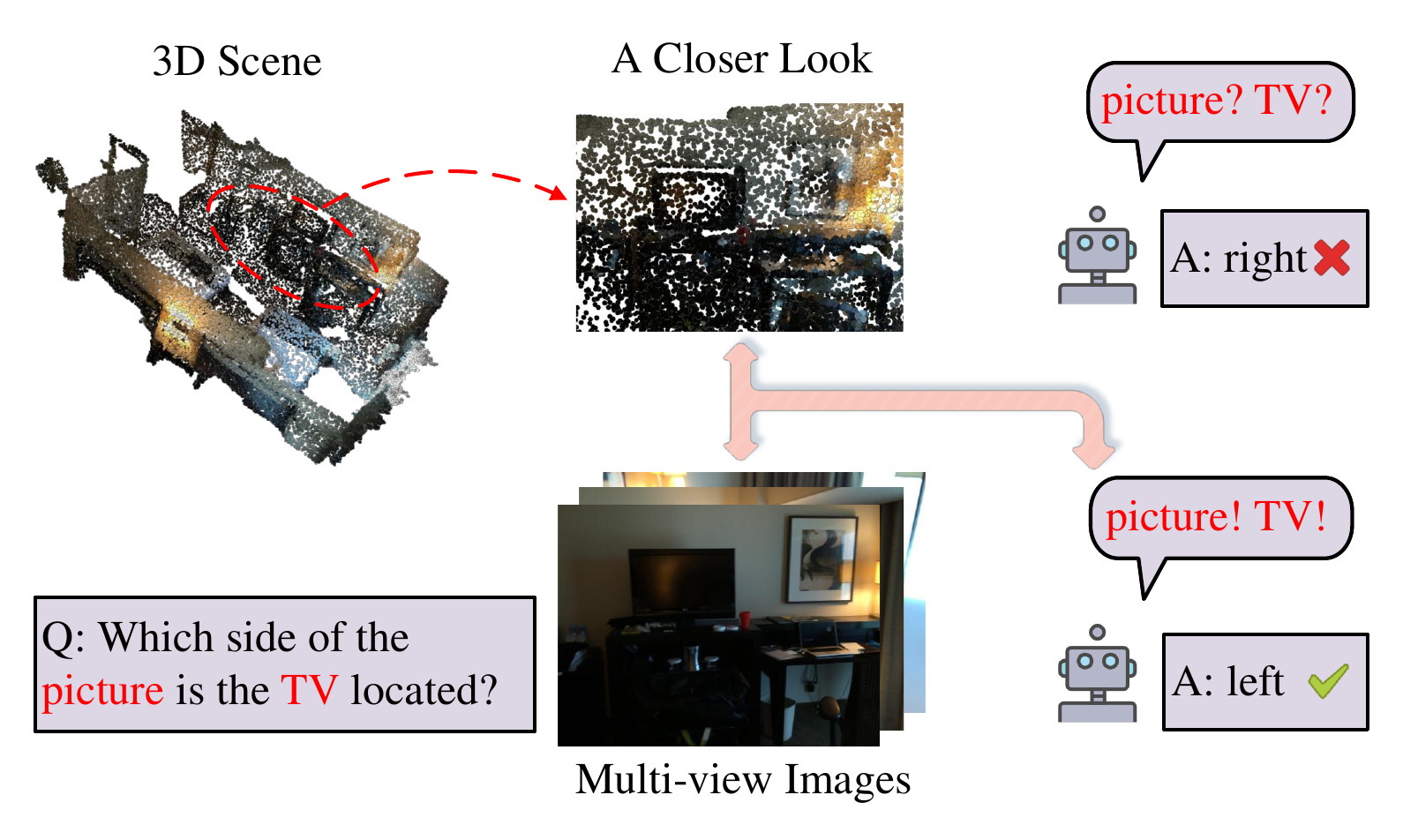}
    \vspace{-20pt}
    \caption{Comprehensive scene perception with dual-vision (point clouds and multi-view images). It is difficult to accurately perceive some flat and small objects (\eg, TV, picture, carpet, phone, etc.) by relying solely on the visual information of point clouds, while multi-view images have richer local texture information and provide more comprehensive visual signals for 3D QA. }
       \vspace{-10pt}
    \label{fig:motivation1}
\end{figure}

Many efforts have been made to address the challenges of 3D QA. For example, ScanQA ~\cite{azuma2022scanqa} introduced a 3D perception-based model to fuse 3D and language information. To capture rich high-level semantic relations among objects, 3DGraphQA~\cite{wu20243dgraphqa} proposed a Graph Transformer-based model for intra-graph and inter-graph feature fusion.  However, most of these methods predominantly rely on 3D point clouds as the primary source of visual information, overlooking the critical role of multi-view images for comprehensive 3D scene perception and reasoning. For example, consider the question given in \cref{fig:motivation1}, ``Which side of the picture is the TV located?" not only requires recognizing entities in geometric scenes but also understanding the complex semantic and spatial relations between scene entities and questions. However, it is difficult for existing 3D QA models to accurately identify some flat and small objects (\eg, TV, picture, carpet, phone, etc.) by relying solely on point cloud information, while multi-view images can make up for this with rich local texture details ~\cite{chen2017mv3d}.

\begin{figure*}
    \centering
    \includegraphics[width=0.9\linewidth]{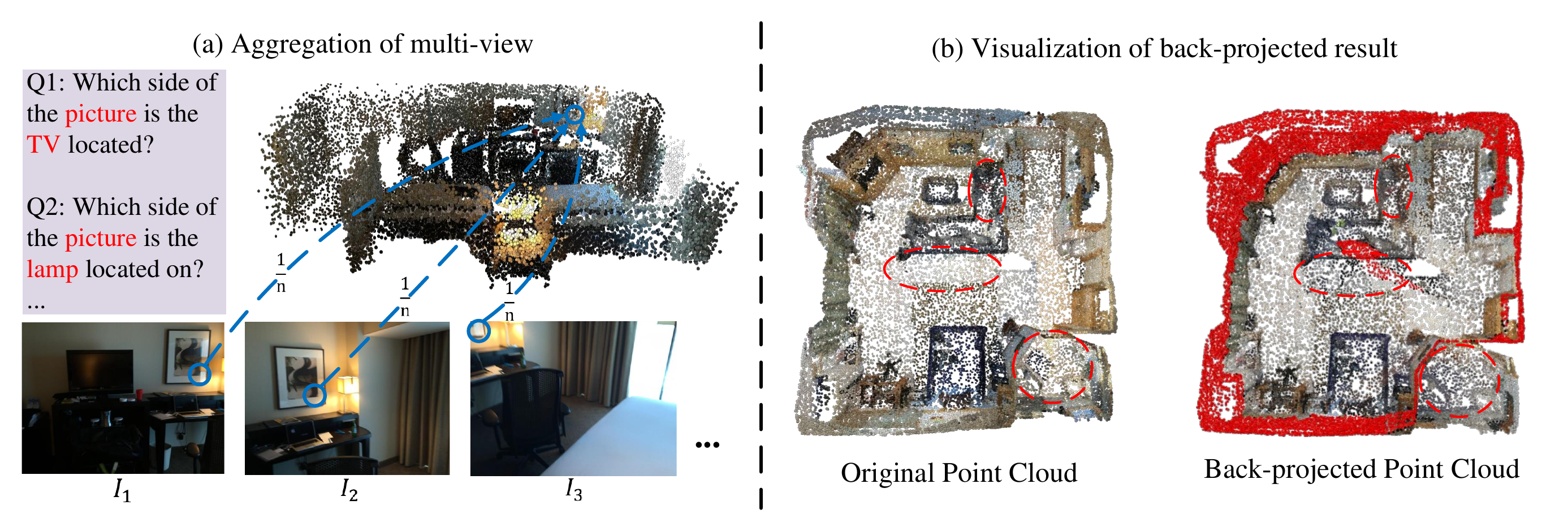}
        \vspace{-15pt}
    \caption{Inherent limitations of back-projection illustrated with a sample from the ScanQA dataset. (a) When aggregating features for coordinates from n mapped multi-view images, each view’s weight remains constant at $\frac{1}{n}$, regardless of the question context. (b) Feature degradation occurs during back-projection from multi-view images to 3D point cloud space due to inherent noise in camera poses, absence of certain views, and complex occlusions, reducing reliability at the edges of the field of view and in occluded areas. Red color points indicate points missed during back-projection (i.e., invalid points), and red ellipses highlight areas with noticeable degradation compared to the original point cloud features.}
       \vspace{-10pt}
    \label{fig:motivation2}
\end{figure*}

To take advantage of multi-view images, a naive approach inspired by 3DMV~\cite{dai20183dmv} is to back-project the multi-view image features into point cloud coordinates, pool the features from multiple views for aggregation, and then simply concatenate them with the point cloud features. However, experiments conducted by ScanQA ~\cite{azuma2022scanqa} demonstrated that this straightforward approach is ineffective for their 3D QA tasks. We believe this is due to the inherent limitations of the back-projection. As shown in \cref{fig:motivation2}(a), the weights of each view remain fixed when aggregating multi-view features, though ideally, the importance of features from different views should vary based on the specific question. Furthermore, as shown in \cref{fig:motivation2}(b), inherent noise in camera poses, the absence of certain views, and complex occlusions lead to unavoidable feature degradation during back-projection from multi-view images to 3D point cloud space. This reduces feature reliability, especially at the edges of the field of view and in occluded regions.

To address the aforementioned issues, we propose DSPNet, a novel dual-vision scene perception network designed to comprehensively integrate multi-view and point cloud features, adaptively fuse visual information, and perform more effective context-guided reasoning for robust 3D QA.
To prioritize view images more closely aligned with the textual content, we introduce a Text-Guided Multi-View Fusion (TGMF) module to integrate back-projected multi-view features by weighting them according to the learnable importance of each image relative to the question. Facing the inherent limitations of feature degradation in back-projection, we design an Adaptive Dual-vision Perception (ADVP) module to adaptively fuse back-projected image features with point cloud features into a unified point-level visual representation by filtering high-confidence point features and suppressing low-confidence point features. To achieve efficient and detailed vision-language interaction, we propose a Multimodal Context-guided Reasoning (MCGR) module with $L$ layers. This module mitigates the high computational cost and feature redundancy of direct cross-modal attention on dense visual features, as well as the semantic loss caused by downsampling, while preserving spatial fidelity and semantic granularity through context-guided reasoning.

Our DSPNet is validated on the ScanQA~\cite{azuma2022scanqa} and SQA3D datasets~\cite{ma2022sqa3d}. The results demonstrate that our DSPNet achieves state-of-the-art performance on these benchmarks. Our main contributions can be summarized as follows:
\begin{itemize}
    \item To achieve comprehensive scene perception and reasoning for 3D QA, we propose a novel Dual-vision Scene Perception Network (DSPNet) based on point clouds and multi-view images. Extensive experiments demonstrate that our DSPNet outperforms all baseline methods on SQA3D and ScanQA datasets.  
    \item We introduce a Text-guided Multi-view Fusion (TGMF) module to integrate multi-view image features, allowing the model to prioritize views that are more closely aligned with the text content.
    \item We design an Adaptive Dual-vision Perception (ADVP) module that adaptively fuses back-projected image features with point cloud features into a unified visual representation, coupled with a Multimodal Context-guided Reasoning (MCGR) module for comprehensive 3D scene reasoning with cross-modal contextual interaction.
\end{itemize}

\section{Related Work}
\label{sec: relate}

\subsection{3D Question Answering}

The current 3D question answering (QA) methods primarily focus on two key task settings~\cite{lei2023recent}: 3D visual question answering (3D VQA)~\cite{azuma2022scanqa} and 3D situated question answering (3D SQA) ~\cite{ma2022sqa3d}. 3D VQA focuses on question answering tasks in complex scenes, requiring the model to have strong spatial perception and reasoning capabilities. In contrast, 3D SQA introduces a contextual description of the agent's position and orientation in the task setting, requiring the agent to perceive scene and answer questions from a first-person perspective. Inspired by 2D VQA models (\eg, MCAN~\cite{yu2019mcan}, GraghVQA~\cite{liang2021graghvqa}), researchers have attempted to design similar architectures in the 3D QA domain to effectively fuse features from 3D point clouds and text.

For 3D VQA, ScanQA~\cite{azuma2022scanqa} 
designed a fusion module composed of Transformer~\cite{vaswani2017transformer} layers and a fusion layer~\cite{yu2019mcan} integrate contextualized word representations of question with object proposal features, followed by a classification layer for answer prediction.
For 3D SQA, SQA3D~\cite{ma2022sqa3d} built upon ScanQA~\cite{azuma2022scanqa} by utilizing a shared-parameter text encoder for additional situation description encoding. It sequentially fuses 3D object proposal features with situation description and question features through Transformer~\cite{vaswani2017transformer}. 
To capture rich high-level semantic relations among objects, 3DGraphQA~\cite{wu20243dgraphqa} proposed a graph-based 3D QA method, which consists of a graph transformer model for intra-graph feature fusion and a bilinear graph neural network for inter-graph feature fusion.



Following the recent success of 2D VLM pre-training in various downstream tasks, researchers have begun exploring the pre-training paradigms in 3D scene understanding. These methods aim to obtain universal 3D vision-language representations through knowledge transfer from 2D VLMs or large-scale data pre-training.
Multi-CLIP~\cite{delitzas2023multiclip} utilized contrastive learning to align 3D scene representations with corresponding text embeddings and multi-view image embeddings in the feature space of CLIP~\cite{radford2021learning} , transferring knowledge from CLIP to enhance 3D vision-language understanding capabilities. 3D-VisTA~\cite{zhu20233d} pre-trained on the large-scale scene-text paired dataset ScanScribe~\cite{zhu20233d}, employing masked language modeling, masked object modeling, and scene-text matching strategies. During fine-tuning, 3D-VisTA efficiently adapted to various downstream tasks by adding lightweight task-specific head structures, without requiring additional auxiliary losses or task-specific optimization techniques.

However, most of the existing methods overlook the significance of multi-view images in comprehensive scene perception and reasoning. To address the existing limitations in multi-view image feature fusion and to enhance comprehensive 3D scene reasoning through cross-modal contextual interaction, we propose a novel Dual-vision Scene Perception Network (DSPNet) for robust 3D question answering.

\subsection{3D Visual Grounding}
3D visual grounding(3D VG)~\cite{chen2020scanrefer,achlioptas2020referit3d} is a 3D language object localization task. ScanRefer~\cite{chen2020scanrefer} proposed a novel approach for 3D object localization using natural language and presents the first large-scale scene-language dataset (i.e., ScanRefer dataset). Subsequently, the ReferIt3D dataset~\cite{achlioptas2020referit3d} is introduced to support fine-grained 3D object identification in real-world scenes, providing detailed multi-instance labels to facilitate distinguishing instances within the same object class. Based on the two datasets, a large amount of research~\cite{cai20223djcg, zhao20213dvg,chen2022language} have been dedicated to the 3D VG tasks. Additionally, these datasets have also been explored for pre-training in 3D QA (\eg, 3D-VisTA~\cite{zhu20233d}, Multi-CLIP~\cite{delitzas2023multiclip}). 3D VG focuses on identifying and localizing specific objects in 3D scenes based on natural language descriptions, while 3D QA extends beyond localization to include spatial reasoning and scene understanding to answer questions about the environment. Although both tasks are multimodal and involve 3D language interaction, 3D QA emphasizes contextual comprehension and relational inference within 3D scenes. 

\subsection{Multi-view Based 3D Perception}
Recently, fusing point cloud and multi-view images in 3D detection has attracted increasing interest. Early works~\cite{dai20183dmv,chen2017mv3d,yoo20203dcvf,liang2018deep} projected 3D queries to multi-view images for collecting useful semantics. While PointPainting~\cite{vora2020pointpainting} and PointAugmenting~\cite{wang2021pointaugmenting} directly decorated raw 3D points with 2D semantics. 3D-CVF~\cite{yoo20203dcvf} performed multi-modal fusion at both point and proposal levels. However, since 3D points are inherently sparse, a hard association approach wastes the dense semantic information in 2D features. Recently, multi-modal 3D detectors ~\cite{wang2024embodiedscan,liang2022bevfusion, jiao2023msmdfusion,li2022unifying} back-projected dense 2D seeds to 3D space for learning the 2D-3D joint representation in a shared space. However, most of them overlook the limitation of feature degradation in back-projection from 2D to 3D. Moreover, they neglect the influence of textual semantics in multi-view feature fusion.

\section{Methodology}
\begin{figure*}
    \centering
    \includegraphics[width=0.98\linewidth]{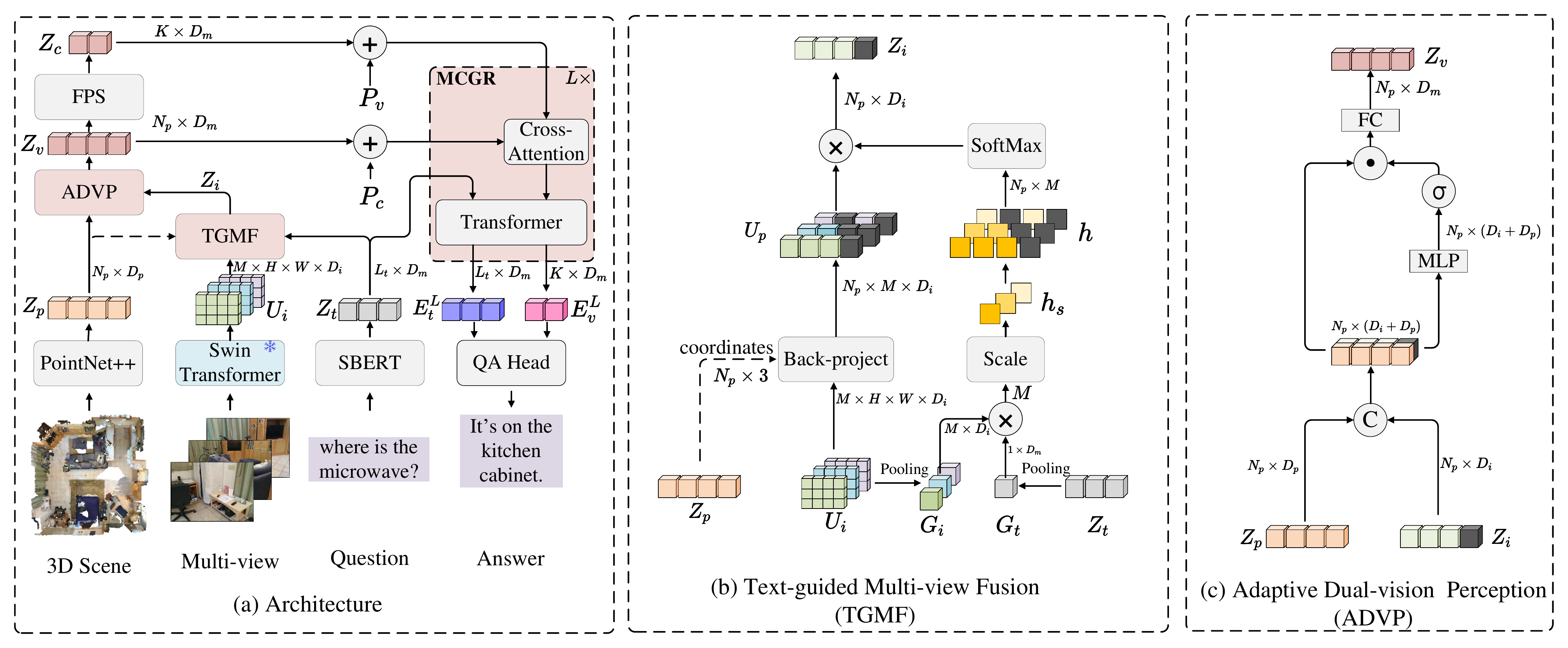}
        \vspace{-10pt}
    \caption{ (a) The overall architecture of the DSPNet: it takes the 3D scene, multi-view images, and question as the inputs, ultimately output answers to questions. (b) The Text-guided Multi-view Fusion (TGMF) module aims to fuse the multi-view features. (c) The Adaptive Dual-vision Perception (ADVP) module aims to adaptively perceive the vision information derived from point cloud and multi-view images.}
       \vspace{-10pt}
    \label{fig:method}
\end{figure*}
In this section, we provide a detailed introduction to our DSPNet, which employs dual-vision comprehensive scene perception to address the task of 3D question answering.
\subsection{Overall Architecture}
 As illustrated in \cref{fig:method}(a), the point cloud $P$ of the 3D scene, the question $T$, and the multi-view images $I$ are served as input for DSPNet, which aims to predict correct answer vector $\alpha \in \mathbb{R}^{N_{\alpha}}$ for the $N_{\alpha}$ answer candidates. DSPNet first encodes the text via a text encoder, encodes the multi-view images via a frozen image encoder, and processes the point cloud via a 3D encoder. Then, we introduce a Text-guided Multi-view Fusion (TGMF) module to fuse the features from multi-view and design an Adaptive Dual-vision Perception (ADVP) module to adaptively perceive the vision information derived from point cloud and multi-view images. Finally, we incorporate a Multimodal Context-guided Reasoning (MCGR) module that facilitates efficient cross-modal interaction between visual and language for comprehensive scene reasoning and question answering.
\subsection{Modal Encoder}
\textbf{3D Encoder.} Given an RGB-colored input point cloud $p \in \mathbb{R}^{N \times 6}$, where $N$ represents the number of points, most prior methods~\cite{ma2022sqa3d,azuma2022scanqa,delitzas2023multiclip} adopt a pre-trained VoteNet~\cite{qi2019votenet} detector to acquire object-level tokens $Z_p \in \mathbb{R} ^ {N_{o} \times D_{o}}$ as the visual representation, where $N_o$ is the number of object proposals, and $D_o$ is the dimension of object-level feature. However, these methods exhibit several limitations: (1) Detection based feature extraction approaches often overlook non-object areas within the scene, which are essential in some reasoning scenarios (\eg, carpets on the floor, pictures on the wall, hanging lamps on the ceiling). (2) After object-level abstraction, high-level information of the scene (\eg, the layout of a bedroom, the corners of a kitchen) is lost in the visual representation. (3) Joint optimization of detection and reasoning tasks requires careful balancing of their respective loss functions, potentially diverting focus from the primary objective of scene reasoning.

In light of these, we adopt a pre-trained PointNet++~\cite{qi2017pointnet++} from VoteNet~\cite{qi2019votenet} (i.e., VoteHead of VoteNet is discarded) as our 3D encoder. Specifically, the 3D encoder comprises several set abstraction layers and feature propagation (upsampling) layers with skip connections. It processes the input point cloud and outputs a subset of points, referred to as seed points. These seed points are represented by their XYZ coordinates and an enriched feature vector $Z_p \in \mathbb{R}^{N_p \times D_p}$, where $N_p$ denotes the number of seed points, and $D_p$ represents the dimension of the point-level features.

\noindent \textbf{Image Encoder.} Given M multi-view images, we employ a pre-trained Swin Transformer~\cite{liu2021swin} to extract multi-view image features $U_i \in \mathbb{R}^{M  \times H \times W \times D_i}$, where $M$ denotes the number of images, $D_i$ represents the feature dimension, and $H \times W$ indicates the spatial resolution of the feature maps.
\noindent \textbf{Text Encoder. }To robustly capture both local and global features of the situation description and question, we adopt a pre-trained Sentence-BERT (SBERT)~\cite{reimers2019sbert} to extract context word-level features $Z_t \in \mathbb{R}^{L_t \times D_m}$, where $L_t$ denotes the sequence length and $D_m$ represents the feature dimension. To unify two task settings of 3D VQA and 3D SQA, in 3D SQA, we directly concatenate the situation description and question as the input of the text encoder as in ~\cite{zhu20233d}.

\subsection{Text-guided Multi-view Fusion}

To fuse M multi-view image features $U_i \in \mathbb{R}^{M  \times H \times W \times D_i}$ from the image encoder, we design a Text-guided Multi-view Fusion (TGMF) module that performs back-projection ~\cite{dai20183dmv} and text-guided fusion to merge these features. 

As illustrated in \cref{fig:method}(b), we initially back-project $U_i$  into 3D coordinates space of $Z_p$ by leveraging the known camera intrinsic and extrinsic parameters associated with each image, obtaining multi-view back-projected features $U_{p} \in \mathbb{R}^{N_p \times M  \times D_i}$ based on point-to-pixel correspondences. Since feature representations of the same entity often vary across different views, especially for entity relations in first-person perspective, we introduce an attention mechanism to learn context-specific importance weights  $s \in \mathbb{R}^{N_p \times M }$ of multi-view for each point location. These weights are then used to preferentially aggregate multi-view information:
\begin{equation}
    Q = G_iW_q, \quad 
    K = G_tW_k, \quad 
    h_s =  \frac{QK^T}{\sqrt{d_k}}
\end{equation}
\begin{equation}
    s = \text{SoftMax}(h, dim=1), \quad Z_i = sU_{p}
\end{equation}
where $G_i \in \mathbb{R}^{M \times D_i}$ is the global pooling feature  of multi-view image features $U_i$,  $G_t \in \mathbb{R}^{1 \times D_m}$ is the global pooling feature of contextualized word-level text features $Z_t$ extracted by the text encoder, $W_{q} \in \mathbb{R}^{D_i \times d_k}$ and $W_{k} \in \mathbb{R}^{D_m \times d_k}$ are learnable weights that project $G_i$ and $G_t$ into a same latent space, $h \in \mathbb{R}^{ N_p \times M }$ is derived by duplicating $h_s \in \mathbb{R}^{M} $ across all valid back-projected points, $Z_i \in \mathbb{R}^{N_p \times D_i}$ denotes the weighted aggregated feature from multi-view back-projected features.
\subsection{Adaptive Dual-vision Perception}

Given the back-projected features $Z_i$  and point cloud features $Z_p$, we aim to adaptively fuse the texture-rich back-projected features with spatially-informative point cloud features into a unified visual representation. 


Inspired by SENet~\cite{hu2018senet}, we design an Adaptive Dual-vision Perception (ADVP) module to point-wise and channel-wise filter high-confidence features and suppress low-confidence ones. As illustrated in \cref{fig:method}(c), after concatenating the back-projected features  $Z_i$ and point cloud features  $Z_p$, we utilize a MLP : $\mathbb{R}^{D_i+D_p} \rightarrow \mathbb{R}^{D_i+D_p}$ and sigmoid function ($\sigma$) to learn the importance of each feature channel at each point. Let $Z_h \in \mathbb{R}^{N_p \times (D_i+D_p)}$ denote the refined features, which are calculated as follows:
\begin{equation}
    Z_h =  \sigma(\text{MLP}([Z_i, Z_p])) \odot [Z_i, Z_p]
\end{equation} 
where $\sigma(\text{MLP}([Z_i, Z_p])) \in \mathbb{R}^{N_p \times (D_i+D_p) }$,  and $\odot$ denotes element-wise multiplication. Finally, a fully connected layers (FC), are employed to map and obtain refined point-level visual features $Z_v \in 
 \mathbb{R}^{N_p \times D_m} $:
\begin{equation}
    Z_v = \text{FC}(Z_h)
\end{equation}

\subsection{Multimodal Context-guided Reasoning}

After acquiring refined point-level visual features $Z_v $ and the contextualized word-level text features $Z_t$ extracted by the text encoder, we aim to derive cross-modal representations through vision-language interaction in the shared semantic space. Therefore, we introduce a Multimodal Context-guided Reasoning (MCGR) module with $L$ layers, which ensures computational efficiency while mitigating the semantic information loss caused by downsampling. 



Initially, we apply farthest point sampling (FPS) to sample $K$ points from the dense point-level visual features $Z_v$, resulting in sparse candidate features $Z_c \in \mathbb{R}^{K \times D_m}$. We then get the position embeddings $P_{v(c)}$ by passing the corresponding coordinates $p_{v(c)}$ through a learnable $\text{MLP}_{v(c)}: \mathbb{R}^{3} \to \mathbb{R}^{D_m}$. The position embeddings are added to $Z_v$ and $Z_c$ to form the dense visual embeddings $E_v$ and sparse visual embeddings $E_c$, respectively:
\begin{equation}
    E_{v(c)} = Z_{v(c)} + \text{MLP}_{v(c)}(p_{v(c)})
\end{equation}

We send $E_{v}$, $E_{c}$ and text features $Z_t \in L_t \times D_m$ to MCGR module, each layer of which contains a cross-attention and transformer sub-layer. Inside the $i$-th layer, the cross-attention sub-layer is first applied. The query are the fused visual output $E^{i-1}_c$ (where $E^{0}_c = E_c$ initially) of the ($i$-1)-th layer, and context vectors are the dense visual embeddings $E_{v}$:
\begin{equation}
    h^{i}_{c} = \text{CrossAtt}(E^{i-1}_c, E_{v})
\end{equation} 
This interactive process can capture essential point features from dense point visual features under context guidance. And then $h^{i}_{c}$ interacts with the fused text feature $E^{i-1}_t$ (where $E^{0}_t = Z_t$ initially) through a transformer sub-layer:
\begin{equation}
    [E^{i}_c, E^{i}_t] = \text{Transformer}([h^{i}_{c}, E^{i-1}_t])
\end{equation}
where $[\cdot,\cdot]$ denotes the concatenation operation along the sequence dimension.


\subsection{Training Objective}
\textbf{Question Answering Head.}  Following~\cite{azuma2022scanqa}, we feed the fused text features output $E^{L}_t \in \mathbb{R}^{ L_t \times D_m} $ and the fused visual features output $E^{L}_v \in \mathbb{R}^{ K \times D_m}$ into a modular co-attention network (MCAN)~\cite{yu2019mcan} to predict answer $\alpha \in \mathbb{R}^{N_{\alpha}}$ for the $N_{\alpha}$ answer candidates.

\begin{table*}
    
    \centering
    \setlength{\tabcolsep}{5pt}
    \begin{tabular}{l c c c c c c c c}
        \toprule
        \multirow{2}{*}{Method} & \multirow{2}{*}{Pre-trained} & \multicolumn{6}{c}{Test set} & \multirow{2}{*}{Avg.} \\
        \cline{3-8}
        ~ & ~& What (1,147) & Is (652) & How (465) &Can (338) &Which (351) &Other (566) & ~\\
        \hline
        ClipBERT~\cite{lei2021clipbert}& ×  &30.2 &60.1 &38.7 &63.3 &42.5 &42.7 &43.3 \\
        MCAN~\cite{yu2019mcan}& × &28.9 &59.7 &44.1 &68.3 &40.7 &40.5 &43.4 \\
        ScanQA~\cite{azuma2022scanqa}& ×  &28.6 &65.0&47.3 &66.3 &43.9 &42.9 &45.3 \\
        SQA3D~\cite{ma2022sqa3d}& × &33.5 &\textbf{66.1} &42.4 &\underline{69.5} &43.0 &46.4 &47.2 \\
        Multi-CLIP~\cite{delitzas2023multiclip}&$\surd$ & - & - & - & - &- &- &48.0 \\
        3D-VisTA~\cite{zhu20233d}& × &32.1 &62.9 &\underline{47.7} &60.7 &45.9 &\underline{48.9} &46.7\\
        3D-VisTA~\cite{zhu20233d}&$\surd$  &34.8 &63.3 &45.4 &\textbf{69.8} & \underline{47.2} &48.1 &48.5\\
        3DGraphQA~\cite{wu20243dgraphqa}& × &\underline{36.4} &64.7&46.1 &\textbf{69.8} &\textbf{47.6} &48.2 &\underline{49.2}\\
        \hline
        DSPNet (ours)& ×& \textbf{38.2}& \underline{66.0}& \textbf{51.2}& 66.6& 42.5& \textbf{51.6}& \textbf{50.4} \\
        \bottomrule
    \end{tabular}
        \vspace{-5pt}
    \caption{The question answering accuracy on the SQA3D dataset. In the test set column: the brackets indicate the number of samples for each type of question. The best results are in bold, and the second-best ones are underlined.}
      \vspace{-10pt}
    \label{tab:sqa}
\end{table*}

\noindent \textbf{3D VQA task.} We model the final loss as a linear combination of answer classification loss $L_{ans}$, object classification loss $L_{cls}$ and reference object center localization loss $L_{loc}$. For $L_{loc}$, the location closes to a ground truth object center (within 0.3 meters) is considered a ground truth location, as in~\cite{qi2019votenet, azuma2022scanqa}. We consider the above three training objectives as multi-label classification problems. Since the labels are noisy (\eg, some samples' answers do not include all the correct answers with different expressions in the candidate set, and some samples do not have the ground truth labels of the reference objects), we use soft-ranked cross entropy loss as the loss function for multi-class classification:
\begin{equation}
L(y, p) = -\log \left( \sum_{i=1}^{N} y_i p_i \right)
\end{equation}
where $y_i \in \{0, 1\}$ is the target label, $p_i \in [0,1]$ is the predicted label confidence through softmax normalization.
The final loss is computed as: 
\begin{equation}
    L_{\text{3DVQA}} = L_{ans} + \lambda_1L_{cls} + \lambda_2L_{loc}
\end{equation}
where $\lambda_1$ and $\lambda_2$ are the weighting factors. In our experiments, we set all these hyper-parameters to 1.

\noindent \textbf{3D SQA task.} We use the answer classification loss for training (i.e., $L_{\text{3DSQA}}=L_{ans}$) as in ~\cite{zhu20233d}.

\section{Experiments}
\label{sec:exp}   
In this section, we validate our DSPNet on two 3D question answering tasks. The tasks for evaluation are (a) 3D situated question answering on the SQA3D dataset~\cite{ma2022sqa3d}  and (b) 3D visual question answering on the ScanQA dataset~\cite{azuma2022scanqa}. 

\subsection{Experimental Setup}
\textbf{Dataset.} The ScanQA dataset~\cite{azuma2022scanqa} contains 41,363 diverse question-answer pairs and 3D object localization annotations for 800 indoor 3D scenes of the ScanNet dataset ~\cite{dai2017scannet}. The SQA3D~\cite{ma2022sqa3d} dataset is designed for embodied scene understanding by integrating situation understanding and situated reasoning. It consists of 6.8k unique situations based on 650 ScanNet scenes, accompanied by 20.4k descriptions and 33.4k diverse reasoning questions for these situations. ScanNet~\cite{dai2017scannet} is a large-scale annotated 3D mesh reconstruction dataset for indoor spaces, where each scene contains the raw RGB-D sequences.

\noindent \textbf{Implemetation Details.} 
We begin by uniformly sampling multi-view images from the original video at a 0.1 ratio for each scene. Subsequently, we select 20 multi-view images with a resolution of $224 \times 224$ as input, utilizing random sampling during training and uniform sampling during inference. Additionally, for each scene, we sample 40,000 points from the raw point cloud as input, using random sampling during training and farthest point sampling during inference. We use the pointnet++ from pre-trained VoteNet~\cite{qi2019votenet}, the pre-trained Swin Transformer~\cite{liu2021swin} and the MPNet-based pre-trained Sentence-BERT (SBERT)~\cite{reimers2019sbert} while training other modules randomly initialized from scratch following end-to-end manners. The $K$ value is set to 256 in the FPS stage preceding the MCGR module, and the hidden size of the MCGR module is set to 768. The network is trained using AdamW~\cite{loshchilov2017adamw} optimizer with $\beta_1$ = 0.9, $\beta_2$ = 0.999 and a weight decay of $1e^{-5}$. We use 4 GPUs with 12 training samples on each to train the model for 12 epochs. The learning rate schedule includes a 500-step warm-up phase, linearly increasing from $5e^{-5}$ to $1e^{-4}$, followed by cosine decay back to $5e^{-5}$. The text encoder use a 0.1× smaller learning rate. We implement DSPNet in Pytorch and train it with NVIDIA GeForce RTX 3090 GPUs.

\begin{table*}
    \centering
    \begin{tabular}{l c c c c c c c}
        \toprule
        Method  &Pre-trained&EM@1 &EM@10 &BLEU-4 &ROUGE &METEOR &CIDEr \\
        \hline
        Image+MCAN~\cite{azuma2022scanqa}&× &22.3 / 20.8 &53.1 / 51.2 &14.3 / 9.7 &31.3 / 29.2 &12.1 / 11.5 &60.4 / 55.6 \\
        ScanRefer+MCAN~\cite{azuma2022scanqa}&×&20.6 / 19.0 &52.4 / 49.7 &7.5 / 7.8 &30.7 / 28.6 &12.0 / 11.4 &57.4 / 53.4  \\
        ScanQA~\cite{azuma2022scanqa}&×&23.5 / 20.9 &56.5 / \underline{54.1} &12.0 / 10.8 &34.3 / 31.1 &13.6 / 12.6 &67.3 / 60.2  \\
        Multi-CLIP~\cite{delitzas2023multiclip}&$\surd$& 24.0 / 21.5 &- / -&12.7 / \underline{12.9} &35.4 / 32.6 &14.0 / 13.4 &68.7 / 63.2  \\
        3D-VisTA~\cite{zhu20233d}&×&25.2 / 20.4 &55.2 / 51.5 &10.5 / 8.7 &35.5 / 29.6 &13.8 / 11.6 &68.6 / 55.7   \\
        3D-VisTA~\cite{zhu20233d}&$\surd$&\textbf{27.0} / \underline{23.0} & 57.9 / 53.5 & \textbf{16.0} / 11.9 & \underline{38.6} / 32.8 & \underline{15.2} / 12.9 &\underline{76.6} / 62.6  \\
        3DGraphQA~\cite{wu20243dgraphqa}&× &25.6 / 22.3 & \underline{58.7} / \textbf{56.1} &15.1 / \underline{12.9} &36.9 / \underline{33.0} &14.7 / \underline{13.6} &74.6 / \underline{62.9}  \\
        \hline
        DSPNet (ours)&×& \underline{26.5} / \textbf{23.8}& \textbf{58.8} / \textbf{56.1} & \underline{15.4} /  \textbf{15.7}& \textbf{39.3} / \textbf{35.1} & \textbf{15.7} / \textbf{14.3}& \textbf{78.1} / \textbf{69.6}  \\
        \bottomrule
    \end{tabular}
            \vspace{-5pt}
    \caption{Answer accuracy on ScanQA. Each entry denotes ``test w/ object'' / ``test w/o object''. The best results are marked bold, and the second-best ones are underlined.}
      \vspace{-10pt}
    \label{tab:scanqa}
\end{table*}


\begin{table}
\setlength{\tabcolsep}{10pt}
    \centering
    \begin{tabular}{c c c c c}
        \toprule
        TGMF &ADVP &MCGR & ScanQA  & SQA3D\\
        \hline
        × &× & × &22.35  &49.33 \\
        $\surd$ &× & × & 22.69 & 49.58\\
        $\surd$ &$\surd$ &× & 22.80& 49.87 \\
        $\surd$ &× &$\surd$  & 23.23& 49.77 \\
        $\surd$ &$\surd$ &$\surd$ &\textbf{23.47} & \textbf{50.36} \\
        \bottomrule
    \end{tabular}
    \vspace{-5pt}
    \caption{Ablation study of components in our method. Conducted on the validation split of the ScanQA dataset and the test split of the SQA3D dataset, using EM@1 as the metric. See \cref{sub: ablation} for a description of each configuration. }
      \vspace{-20pt}
    \label{tab:ablation1}
\end{table}

\noindent \textbf{Evaluation Metrics.} On the ScanQA dataset~\cite{azuma2022scanqa}, we employ the same evaluation metrics as ~\cite{azuma2022scanqa}, which include EM@1 and EM@10, where EM denotes the exact match and EM@K represents the percentage of predictions that exactly match any ground truth answer among the top K predicted answers. Meanwhile, we utilize BLEU-4, ROUGE, METEOR, and CIDEr as the sentence-level evaluation metrics. On the SQA3D dataset~\cite{ma2022sqa3d}, we adopt the answer accuracy under different types of questions.

\subsection{Results on SQA3D Dataset}
\textbf{Baseline.} We perform a comparison evaluation with several representative baselines on the SQA3D dataset. In particular, we evaluate against ClipBERT~\cite{lei2021clipbert} and MCAN~\cite{yu2019mcan} which are, as reported in prior work ~\cite{ma2022sqa3d} baselines focused on egocentric video and bird-eye view (BEV) image QA.  ScanQA~\cite{azuma2022scanqa} represents a 3D QA baseline that ignores the situational input. SQA3D ~\cite{ma2022sqa3d}  allows location descriptions and questions to interact separately with object proposal features. 
Multi-CLIP~\cite{delitzas2023multiclip} and 3D-VisTA~\cite{zhu20233d} are pre-trained on external 3D-Text paired datasets before being fine-tuned on this dataset. 
3DGraphQA ~\cite{wu20243dgraphqa} is trained on SQA3D Pro dataset by incorporating multi-view images to complement the image information of the first-person perspective situations in the SQA3D dataset.

\noindent \textbf{Results Analysis.} 
Our DSPNet leverages dual-vision scene perception and reasoning, utilizing multi-view images that contain rich local texture details to achieve a more nuanced understanding of scene intricacies. As shown in \cref{tab:sqa}, we achieve the best results on \textit{What}, \textit{How} and \textit{Other} questions and outperforms other methods including those pre-trained on external 3D-Text paired datasets in terms of average accuracy. This validates that our DSPNet has competitive question reasoning capability. In contrast, SQA3D, 3D-VisTA, and 3DGraphQA exhibit better performance on simpler questions with fewer answer options, such as \textit{Is}, \textit{Can}, and \textit{Which}, where answers can often be inferred correctly based on only the question without relying on 3D scenes.  However, these methods exhibit limited capabilities in fine-grained perception and reasoning when answering complex and open-ended questions like \textit{What} and \textit{How}. These validate that our DSPNet can comprehensively understand the 3D scene and infer the correct answer.


\subsection{Results on ScanQA Dataset}

\textbf{Baseline.} We further perform a comparison evaluation with several representative baselines on the ScanQA dataset. 
In particular, we compare with 2D image VQA MCAN~\cite{yu2019mcan} based baselines ~\cite{azuma2022scanqa}, ScanQA~\cite{azuma2022scanqa}, Multi-CLIP~\cite{delitzas2023multiclip}, 3D-VisTA~\cite{zhu20233d} and 3DGraphQA~\cite{wu20243dgraphqa}. 

\noindent \textbf{Results Analysis.} In \cref{tab:scanqa}, our method outperforms existing representative approaches on most evaluation metrics, especially in CIDEr, ROUGE and METEOR, where it significantly surpasses other methods. Specifically, our method's high CIDEr score reflects its effective capture of relevant semantic content in the answers, the elevated ROUGE score indicates comprehensive coverage of key information, and the high METEOR score demonstrates close alignment with reference answers in both vocabulary and structure. Additionally, on the EM@1 metric, our method requires no additional pre-training but achieves performance comparable to 3D-VisTA, which is pre-trained on a external large-scale 3D-Text paired dataset.

\begin{figure*}
    \centering
    \includegraphics[width=0.98\linewidth]{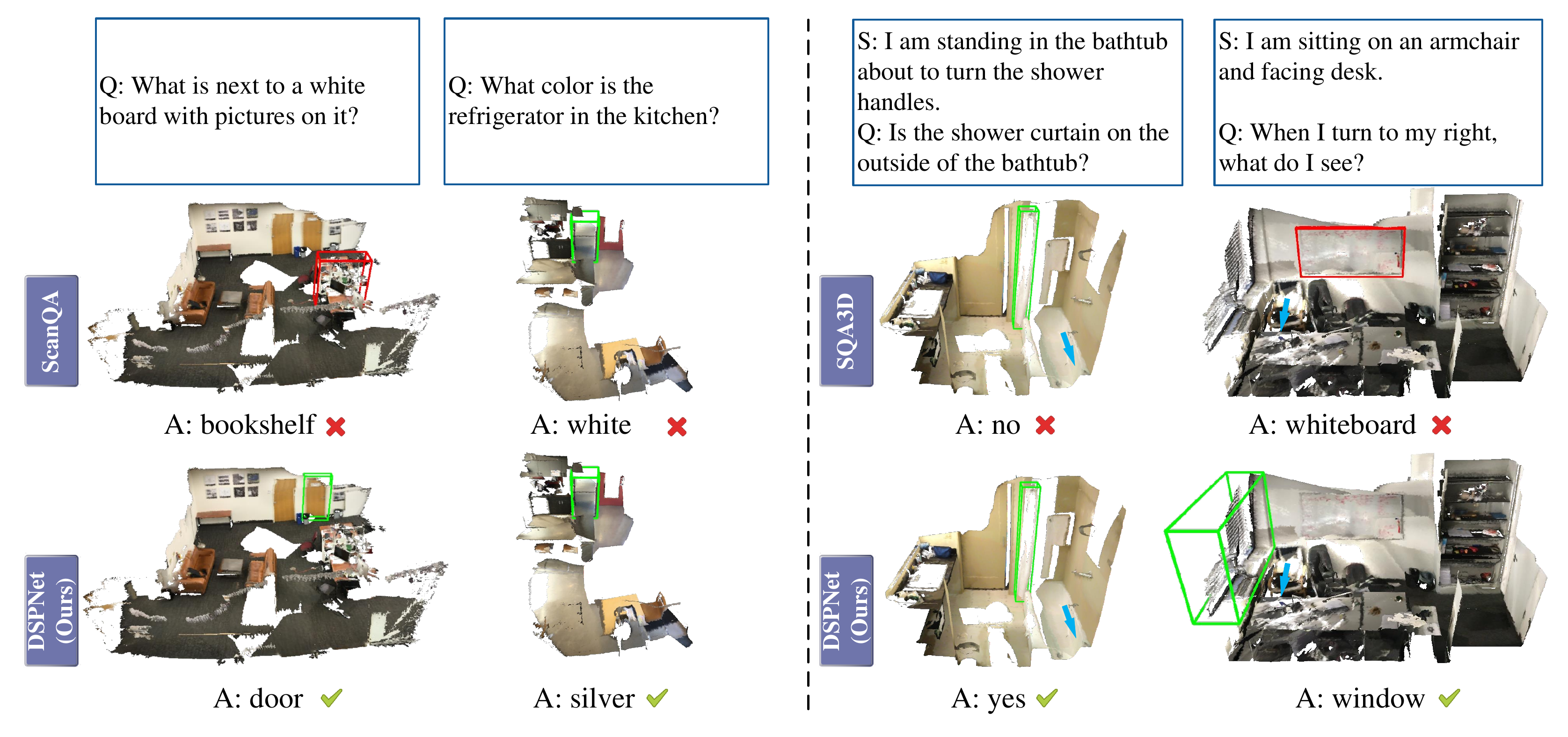}
            \vspace{-10pt}
    \caption{The qualitative comparison of our method with ScanQA and SQA. Our method achieves higher answer accuracy for questions that directly or indirectly involve some challenging entities, such as those with flat shapes and rich local texture details.}
\label{fig:qualitative_study}
\end{figure*}

\subsection{Ablation Studies}
\label{sub: ablation}
We conducted ablation studies on the validation split of the ScanQA dataset and the test split of the SQA3D dataset to evaluate the effectiveness of our proposed components. Starting from a baseline model without any of our modules, we incrementally added the Text-guided Multi-view Fusion (TGMF), Adaptive Dual-vision Perception (ADVP), and Multimodal Context-guided Reasoning (MCGR) modules to assess their individual and combined contributions.
\noindent \textbf{Baseline.} The baseline model excludes TGMF, ADVP, and MCGR. It uses average pooling to aggregate back-projected multi-view features and employs simple concatenation to combine dual visual features. In the reasoning process, it removes the cross-attention sub-layer, performing only basic interactions between text features and candidate visual features sampled from dense point-level features. As shown in \cref{tab:ablation1}, the baseline achieves EM@1 scores of 22.35\% on ScanQA and 49.33\% on SQA3D.

\begin{table}
\setlength{\tabcolsep}{10pt}
    \centering
    \begin{tabular}{l c c}
        \toprule
        Methods &  ScanQA &SQA3D\\
        \hline
        w/o 2D & 22.26& 49.05\\
        DSPNet (ours) &\textbf{23.47} & \textbf{50.36} \\
        \bottomrule
    \end{tabular}
    \vspace{-5pt}
    \caption{Ablation study on the effectiveness of using 2D modality. }
      \vspace{-10pt}
    \label{tab:without 2D}
\end{table}

 \noindent \textbf{Text-guided Multi-view Fusion.} Incorporating the TGMF module improves the baseline performance to 22.69\% on ScanQA and 49.58\% on SQA3D. This indicates that the TGMF, which prioritizes view images aligned with the textual content, is essential for multi-view feature fusion and contributes to overall QA performance.

\noindent \textbf{Adaptive Dual-vision Perception.} After building upon the model with TGMF, adding ADVP further improves performance to 22.80\% on ScanQA and 49.87\% on SQA3D. The results demonstrate the effectiveness of the ADVP module in adaptively fusing back-projected image features with point cloud features.

\noindent \textbf{Multimodal Context-guided Reasoning.} Alternatively, adding MCGR to the model with TGMF improves performance to 23.23\% on ScanQA and 49.77\% on SQA3D. This emphasizes the importance of enhanced multimodal reasoning provided by the MCGR module, as it incorporates dense visual features into contextual interactions via a cross-attention mechanism, significantly preserving scene information and enhancing the model's contextual reasoning capabilities.

\noindent \textbf{Full Model.} Combining all three modules, TGMF, ADVP, and MCGR, into our full model yielded the highest performance, achieving EM@1 scores of \textbf{23.47\%} on ScanQA and \textbf{50.36\%} on SQA3D. The results confirm that the refined features from TGMF and ADVP enhance MCGR's contextual reasoning, which in turn more effectively utilizes these integrated features, leading to optimal overall performance.

\begin{table}[!t]
    \centering
    \begin{minipage}{0.2\textwidth}
        \centering
        \caption*{(a) Number of Views}
        \vspace{-10pt}
        \scalebox{0.90}{
        \begin{tabular}{ccc}
            \toprule
            & ScanQA & SQA3D  \\
            \midrule
            10 & 22.87 & 49.73  \\
            15 & 23.04 & 49.99  \\
            \rowcolor[HTML]{E3E3E3} 20 &\textbf{23.47} & \textbf{50.36} \\
            
            \bottomrule
        \end{tabular}
        }
    \end{minipage}%
    \hfill
    \begin{minipage}{0.2\textwidth}
        \centering
        \caption*{(b) Depth of MCGR}
               \vspace{-10pt}
        \scalebox{0.90}{
        \begin{tabular}{ccc}
            \toprule
            & ScanQA & SQA3D \\
            \midrule
            2 &  23.04& 49.79 \\
            \rowcolor[HTML]{E3E3E3} 4 & \textbf{23.47} & \textbf{50.36} \\
            6 & 22.48 & 49.30 \\
            \bottomrule
        \end{tabular}
        }
    \end{minipage}
    \caption{Ablation study of various design choices. Our settings are marked in gray.}
      \vspace{-10pt}
    \label{tab:ablation2}
\end{table}

\noindent \textbf{Architectural Design}. 
We compare our full model with a variant, ``w/o 2D'', which removes all multi-view images and only adopts the 3D point cloud as visual information input. \cref{tab:without 2D} shows that incorporating local texture details from multi-view images for 3D scene perception and reasoning can bring large improvements to both 3D QA tasks, which proves the necessity of dual vision in our method. In addition, we find that the number of views affects the performance on both 3D QA datasets, especially on ScanQA. As shown in \cref{tab:ablation2}(a), the more views incorporated, the better the performance, as additional views provide richer scene features. We also study the effect of the depth of Multimodal Context-guided Reasoning module by varying the number of layers. As shown in \cref{tab:ablation2}(b), using 4 layers achieves the best performance and simply adding more layers does not help. This is because under the condition of limited-scale 3D QA datasets, deeper networks have stronger representation capabilities but are also more prone to overfitting, so a balance needs to be struck here. 

More ablation studies and analyses are provided in the supplementary material.

\subsection{Qualitative Analysis}
We qualitatively compare our method with ScanQA and SQA3D on the 3D VQA and 3D SQA tasks, respectively.  As shown in \cref{fig:qualitative_study}, our DSPNet performs well in perceiving and reasoning about some challenging entities, such as those with flat shapes and rich local texture details that are difficult to identify based on point cloud geometry alone. Furthermore, DSPNet can distinguish subtle color differences, such as between white and silver, thus enhancing its robustness in identifying fine-grained visual distinctions.

\section{Conclusion and Limitation}
\label{sub:conclusion}
In this paper, we propose DSPNet, a dual-vision network for 3D QA. DSPNet integrates multi-view image features via a Text-guided Multi-view Fusion module. It adaptively fuses image and point cloud features into a unified representation using an Adaptive Dual-vision Perception module. Finally, a Multimodal Context-guided Reasoning module is introduced for comprehensive 3D scene reasoning. Experimental results have demonstrated that DSPNet outperforms existing methods with better alignment and closer semantic structure between predicted and reference answers.

A limitation of DSPNet is that it relies on pre-scanned point clouds and pre-captured multi-view images, which may limit its applicability in dynamic environments. 

{
    \small
    \bibliographystyle{ieeenat_fullname}
    \bibliography{main}
}

\clearpage
\setcounter{page}{1}
\maketitlesupplementary
\begin{table*}[h]
    \centering
    \setlength{\tabcolsep}{5pt}
    \scalebox{1.0}{
    \begin{tabular}{l c c c c c}
        \toprule
        Method & Pre-trained & LLMs-based &Extra dataset & ScanQA & SQA3D  \\
        \hline
        LM4Vision ~\cite{pang2024frozen}& × &$\surd$ & × & - / - &48.1 \\
        PQ3D ~\cite{zhu2024unifying}& ×  & × &$\surd$ &26.1 / 20.0 & 47.1 \\
        GPS ~\cite{jia2024sceneverse}&$\surd$ & × &$\surd$ & 25.0 / 23.5 & 49.9 \\
        LEO~\cite{huang2023leo}&$\surd$ &$\surd$ &$\surd$ & - / - &50.0 \\
        \hline
        DSPNet (Ours)& × &×&×  &\textbf{26.5} / \textbf{23.8} &\textbf{50.4} \\
        \bottomrule
    \end{tabular}
    }
        \vspace{-5pt}
    \caption{The QA accuracy (EM@1) on the ``test w/ object'' / ``test w/o object'' split of ScanQA and the test split of SQA3D. }
      \vspace{-10pt}
    \label{tab:sup-more-baselines}
\end{table*}

\section{More Compared Methods}
\label{sec:more_compared}

We further compare DSPNet with additional state-of-the-art methods that incorporate 3D-language alignment pre-training, external datasets, or large language models (LLMs). As summarized in \cref{tab:sup-more-baselines}, despite not leveraging any of these auxiliary enhancements, DSPNet achieves highly competitive performance on both ScanQA and SQA3D datasets. This demonstrates the effectiveness of our approach, highlighting its capability to perform well without relying on extensive pre-training or external resources.

\section{Results on ``3DQA'' dataset}
We have previously evaluated our method on ScanQA and SQA3D, two widely recognized 3D question answering (3D QA) benchmarks that encompass diverse reasoning tasks, including spatial attribute recognition, embodied activities, navigation, common sense reasoning, and multi-hop reasoning. To further assess the generalizability of our approach, we conduct additional experiments on another 3D QA benchmark introduced by Ye et al.~\cite{ye20243DQA}, named ``3DQA'', which is a human-annotated free-form dataset. For fair comparison with our method, we fine-tune 3D-VisTA~\cite{zhu20233d} from scratch on the ``3DQA'' dataset. As shown in \cref{tab:3dqa_results}, Our method achieves EM@1 scores of 52.0\%, outperforming 3D-VisTA (49.3\%), demonstrating its effectiveness across different 3D QA benchmarks.

\begin{table}[h]
    \centering
    \label{tab:3dqa_results}
    \begin{tabular}{l|cc}
        \toprule
        Method & EM@1 & EM@10 \\
        \midrule
        3D-VisTA~\cite{zhu20233d}           & 49.3 & 88.6  \\
        DSPNet (Ours)      & \textbf{52.0} & \textbf{90.5}  \\
        \bottomrule
    \end{tabular}
    \caption{The question answering accuracy on the validation split of ``3DQA'' dataset.}
\end{table}


\section{More Ablation Studies}
\label{sec:more ablation}
Here we provide more ablation studies on our model.

\noindent \textbf{3D Encoder.}
To evaluate the impact of different pre-trained 3D encoders on our model's performance, we experimented with VoteNet~\cite{qi2019votenet} and PointNet++~\cite{qi2017pointnet++}. PointNet++ extracts local geometric features by hierarchically partitioning point clouds into nested regions and recursively processing them into dense point-level visual features, without the utilization of an explicit object detection module. VoteNet, on the other hand, builds upon PointNet++ by introducing a voting mechanism and a detection head to perform 3D object detection within the point cloud. It generates object proposals by aggregating votes from dense point-level visual features and refines them to localize and classify objects. In our experiments, PointNet++ is initialized from the pre-trained VoteNet, which has been pre-trained on a 3D object detection task in ScanNet~\cite{dai2017scannet} dataset. In the VoteNet configuration, we input object-level visual features from object proposals into our Multimodal Context-guided Reasoning module, rather than using sparse candidate point-level visual features that are sampled from dense point-level visual features. As shown in \cref{tab:more ablation1}, PointNet++ outperforms VoteNet, achieving higher accuracy on both the ScanQA~\cite{azuma2022scanqa} and SQA3D~\cite{ma2022sqa3d} datasets. This suggests that using a 3D encoder without an object detection head enhances the model's generalization ability in 3D QA tasks. The absence of an object detector allows the encoder to learn more generalized and holistic scene features, rather than focusing on specific object categories.

\noindent \textbf{Image Encoder.}
To investigate the impact of different pre-trained image encoders on our model's performance, we conducted experiments with Vision Transformer (ViT)~\cite{dosovitskiy2021vit}, BEiT~\cite{bao2022beit} and Swin Transformer~\cite{liu2021swin}. ViT directly applies a pure transformer structure by splitting images into fixed-size patches and processing them sequentially. BEiT employs a masked image modeling strategy for self-supervised pre-training, learning visual representations through predicting masked image patches.  Swin Transformer introduces a hierarchical architecture with shifted windows for computing self-attention, which efficiently handles various image resolutions. For fair comparison, all experiments are conducted using the base size of these models. As shown in \cref{tab:more ablation2}, Swin Transformer consistently outperforms other architectures, achieving the best performance. These results suggest that advanced image encoders can bolster a model's capabilities in 3D QA tasks, primarily due to their enhanced extraction of multi-view image features that deepen the perception of local texture details within 3D scenes.

\noindent \textbf{Text Encoder.}
To evaluate the effectiveness of different pre-trained text encoders, we experimented with BERT~\cite{devlin2018bert}, RoBERTa~\cite{liu2019roberta} and Sentence-BERT (SBERT)~\cite{reimers2019sbert} architectures. BERT utilizes bidirectional training and masked language modeling to learn contextual representations. RoBERTa builds upon BERT by implementing optimized training strategies, such as extended training duration, increased batch sizes, removal of the next sentence prediction task, and dynamic masking. SBERT leverages siamese network structure to generate semantically meaningful sentence embeddings. In our experiments, we utilize the base size of each model for fair comparison. According to the results presented in \cref{tab:more ablation3}, SBERT delivers the most notable performance enhancements. This improvement highlights the benefit of adopting a powerful text encoder, which helps to gain a deeper understanding of situation descriptions and questions through its strong semantic understanding at the sentence level, significantly improving the model's performance in 3D QA tasks. 

\noindent \textbf{Inference Speed Analysis.}
We conducted an inference speed analysis by measuring the average processing time per sample for different settings of the number of image views. The results indicate that processing time per sample scales with the number of image views, increasing from \textbf{117 ms for 10 views to 171 ms for 15 views and 217 ms for 20 views}. These results demonstrate the significant impact of the number of image views on inference time, highlighting the importance of optimizing scene understanding with fewer multi-view images, which is a promising direction for future research.

 \begin{table}[!h]
    \centering
    \begin{tabular}{ l | cc}
        \toprule
         Encoder & ScanQA & SQA3D  \\
        \midrule
        VoteNet~\cite{qi2019votenet} &22.65  &  49.84\\
        PointNet++~\cite{qi2017pointnet++}& \textbf{23.47} & \textbf{50.36} \\
        \bottomrule
    \end{tabular}
    \caption{Ablation study of different 3D encoders. Conducted on the validation split of the ScanQA dataset and the test split of the SQA3D dataset, using EM@1 as the metric.}
    \label{tab:more ablation1}
    
\end{table}
 \begin{table}[!h]
    \centering
    \begin{tabular}{ l | cc}
        \toprule
         Encoder & ScanQA & SQA3D  \\
        \midrule
        ViT~\cite{dosovitskiy2021vit} &22.46  &  49.39\\
        BEiT~\cite{bao2022beit} &22.63  &  49.87\\
        Swin Transformer~\cite{liu2021swin}& \textbf{23.47} & \textbf{50.36} \\
        \bottomrule
    \end{tabular}
    \caption{Ablation study of different image encoders. Conducted on the validation split of the ScanQA dataset and the test split of the SQA3D dataset, using EM@1 as the metric.}
    \label{tab:more ablation2}
\end{table}

 \begin{table}[!h]
    \centering
    \begin{tabular}{ l | cc}
        \toprule
         Encoder & ScanQA & SQA3D  \\
        \midrule
        BERT~\cite{devlin2018bert} &22.57  &48.68   \\
        RoBERTa~\cite{liu2019roberta} &23.22  &49.47   \\
         SBERT~\cite{reimers2019sbert} &\textbf{23.47} & \textbf{50.36} \\
        \bottomrule
    \end{tabular}
    \caption{Ablation study of different text encoders. Conducted on the validation split of the ScanQA dataset and the test split of the SQA3D dataset, using EM@1 as the metric.}
    \vspace{-10pt}
    \label{tab:more ablation3}
\end{table}

\section{More Qualitative Results.}
\begin{figure}[h]
    \centering
\includegraphics[width=0.48\textwidth]{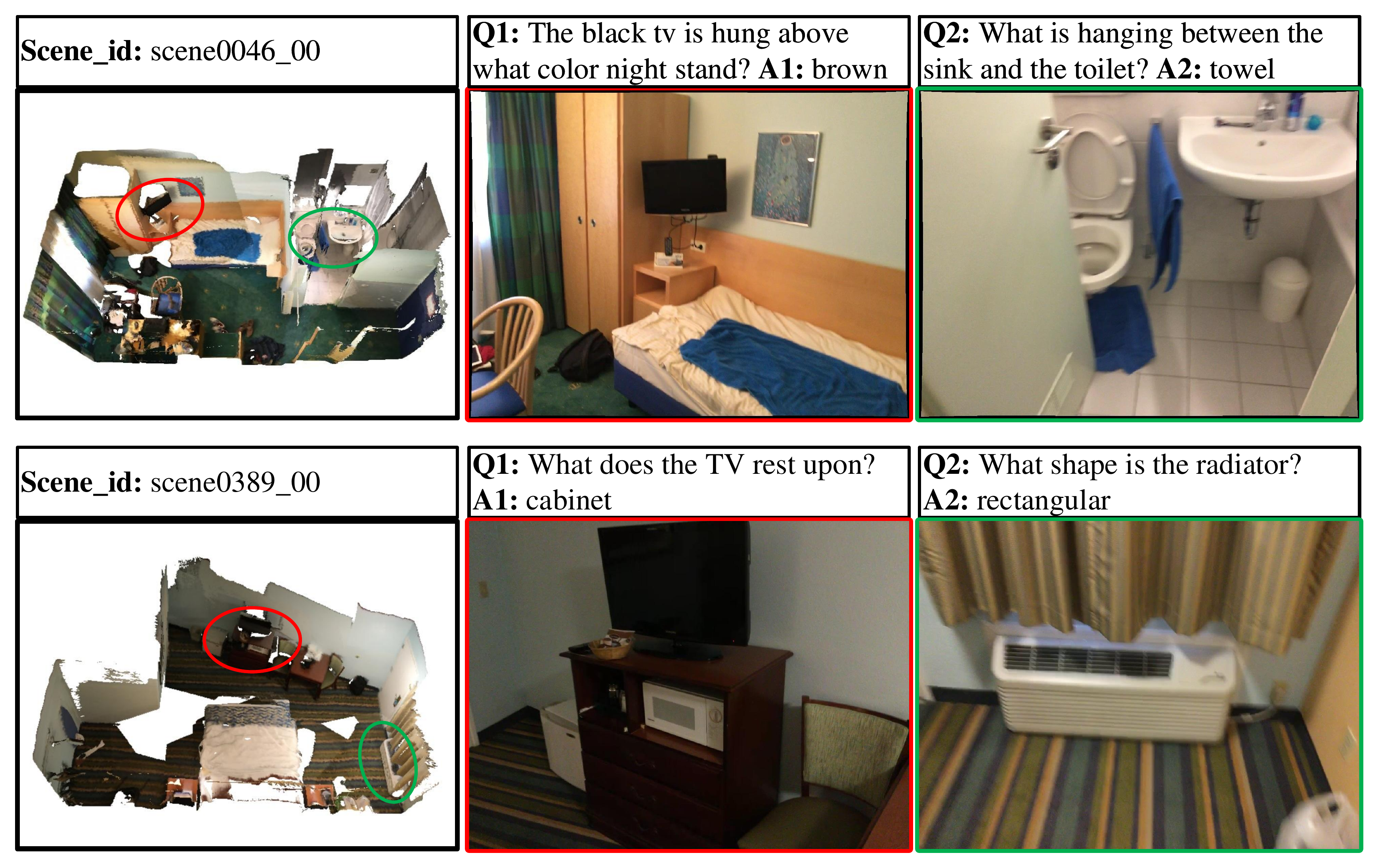}
    \vspace{-20pt}
    \caption{The TGMF module dynamically prioritizes different views according to question context within the same scene.}
    \vspace{-15pt}
    \label{fig:qualitative_results}

\end{figure}

\noindent \textbf{Qualitative Results of TGMF module.} 
We visualized the intermediate results of the TGMF module in \cref{fig:qualitative_results} to provide a clearer understanding of its functionality. Specifically, we showed the image that exhibits the highest context-specific importance weights. From the results, we can see that our TGMF module performs its intended function well.

\begin{figure*}
    \centering
    \includegraphics[width=1.0\linewidth]{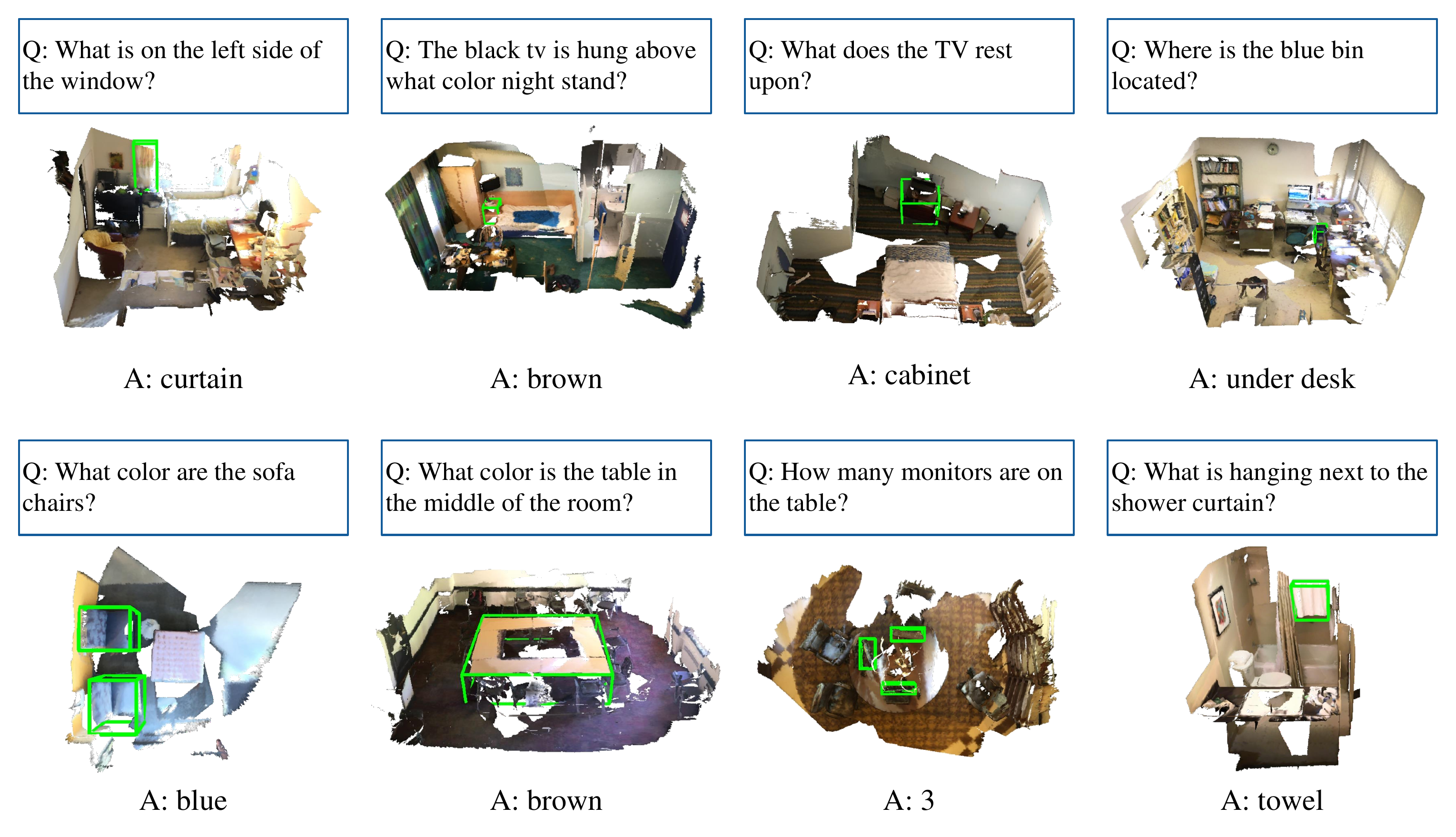}
    \vspace{-15pt}
    \caption{We present more qualitative results on ScanQA dataset.}
       \vspace{-10pt}
    \label{fig:more_qualitative_study_scanqa}
\end{figure*}

\begin{figure*}
    \centering
    \includegraphics[width=1.0\linewidth]{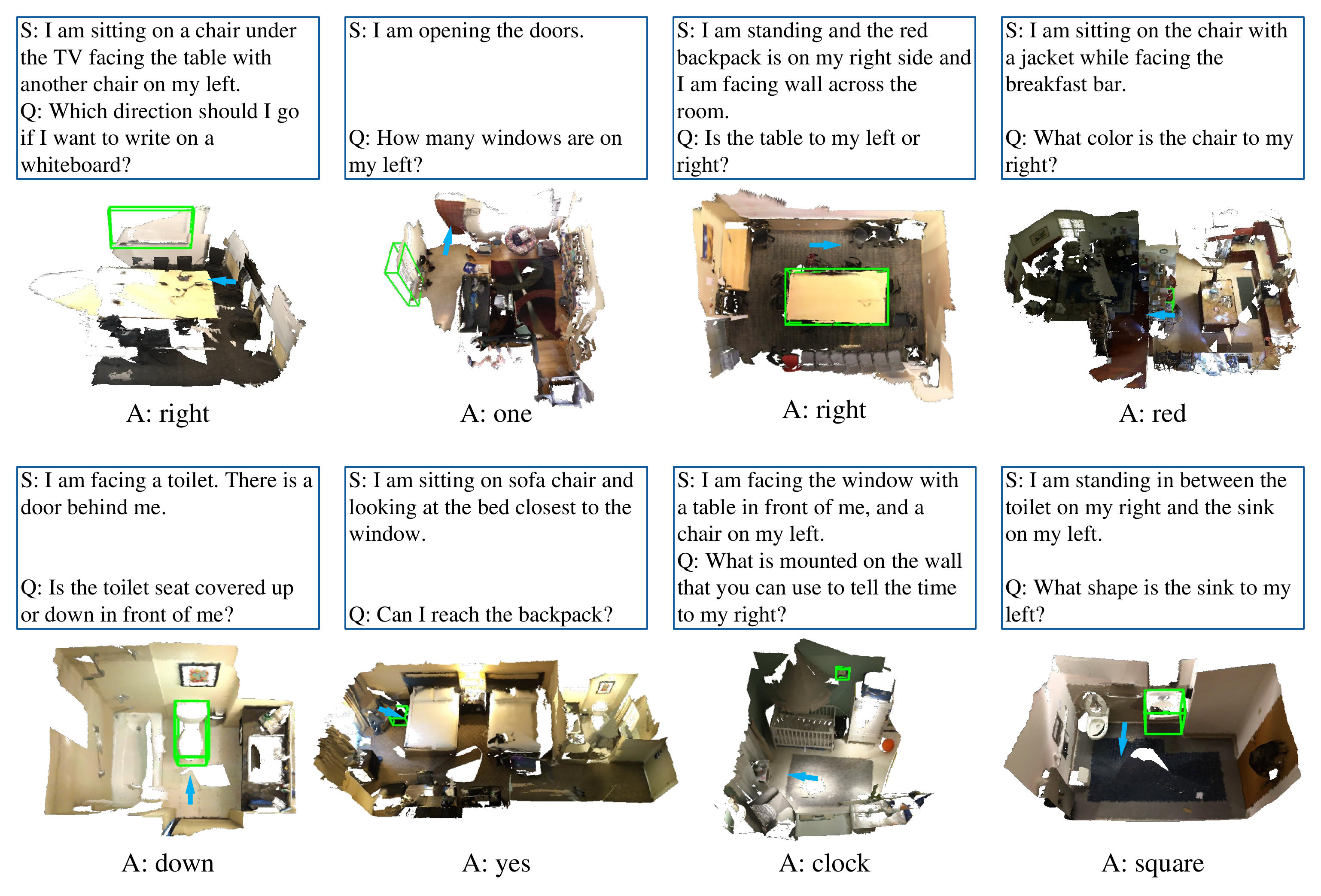}
    \vspace{-15pt}
    \caption{We present more qualitative results on SQA3D dataset.}
    \vspace{-20pt}
    \label{fig:more_qualitative_study_sqa}
\end{figure*}

\noindent \textbf{Qualitative Results of our model.} Additional qualitative results demonstrating the performance of our model are provided in \cref{fig:more_qualitative_study_scanqa} and \cref{fig:more_qualitative_study_sqa}. These results illustrate our model's ability to handle a diverse range of tasks, including querying the locations of objects, identifying characteristics and states of specific objects, counting the number of objects within a scene, and responding to yes/no questions that require commonsense reasoning. From these results, we observe that our method remains robust in complex scenes, despite the varied shapes of the objects involved in the reasoning process, including objects with flat shapes (\eg, whiteboard, TV, clock) and even objects with flexible shapes (\eg, curtain, towel, jacket).

\section{Future Work} 
\label{sec: future} 

\textbf{Adapting to Dynamic Environments.} In future developments of DSPNet, we plan to extend the model's functionality in dynamic environments where changes occur in real-time. This improvement requires evolving our framework to accommodate real-time data acquisition and processing, reducing the dependency on pre-scanned point clouds and pre-captured multi-view images. Such advancements will involve integrating adaptive streaming algorithms that can handle continuous input from moving cameras and sensors.

\noindent \textbf{Multi-modal Alignment.} Further, we intend to enhance DSPNet's ability to perceive and reason within 3D scenes comprehensively through multi-modal integration. We will investigate the alignment of pre-training across 3D scenes, multi-view images, and text related to scenes. This effort will focus on developing a comprehensive multi-modal pre-training approach that utilizes the inherent relationships among these modalities. By applying strategies like contrastive learning and cross-modal distillation, we aim to improve the semantic consistency and contextual understanding across visual and textual data.

\noindent \textbf{Integration with Large Models.} In this paper, we haven't adopt large models due to the limited size of available 3D QA datasets, which restricts the effective training and generalization capabilities of such models. Large models usually require large-scale datasets to avoid overfitting and fully utilize their capacity. In addition, the computational limitations of current devices make it challenging to deploy large models.
However, with the improvement of computing power of modern hardware and the emergence of larger 3D QA datasets in the future, exploring large 3D QA models with dual-vision becomes a promising direction. Our future research will focus on developing scalable architectures to effectively utilize expanded datasets and enhance the model's ability to comprehensively perceive and reason in 3D scenes.

\end{document}